\documentclass{llncs}
\usepackage{pgf}
\usepackage{graphicx}
\graphicspath{ {images/} }
\begin{document}

\title{
TextRay: Mining Clinical Reports to Gain a Broad Understanding of Chest X-rays
}
\titlerunning{TextRay}  
%
\author{Jonathan Laserson\inst{1} \and 
Christine Dan Lantsman\inst{2} \and 
Michal Cohen-Sfady \inst{1} \and 
Itamar Tamir\inst{3} \and
Eli Goz\inst{1} \and 
Chen Brestel\inst{1} \and
Shir Bar\inst{4} \and 
Maya Atar\inst{5} \and 
Eldad Elnekave\inst{1}}
%
%
%
\institute{Zebra Medical Vision LTD, Shefayim, Israel \email{jonil@zebra-med.com} \and
Sheba Medical Center and Tel Aviv University, Israel \and 
Rabin Medical Center, Israel \and 
Technion, Israel Institute of Technology \and 
Ben Gurion University, Israel}

\maketitle        

\begin{abstract}
The chest X-ray (CXR) is by far the most commonly performed radiological examination for screening and diagnosis of many cardiac and pulmonary diseases. There is an immense world-wide shortage of physicians capable of providing rapid and accurate interpretation of this study. A radiologist-driven analysis of over two million CXR reports generated an ontology including the 40 most prevalent pathologies on CXR. By manually tagging a relatively small set of sentences, we were able to construct a training set of 959k studies. A deep learning model was trained to predict the findings given the patient frontal and lateral scans. For 12 of the findings we compare the model performance against a team of radiologists and show that in most cases the radiologists agree on average more with the algorithm than with each other.  

\end{abstract}
\section{Introduction}
Chext X-rays (CXR’s) are the most commonly performed radiology examination world-wide, with over 150 million obtained annually in the United States alone.  CXR’s are a cornerstone of acute triage as well as longitudinal surveillance. Despite the ubiquity of the exam and its apparent technical simplicity, the chest x ray is widely regarded among radiologists as among the most difficult to master\cite{Robinson1999}.

Due to a shortage in supply of radiologists, radiographic technicians are increasingly called upon to provide preliminary interpretations, particularly in Europe and Africa. In the US, non-radiology physicians often provide preliminary or definitive readings of CXRs, decreasing the waiting interval at the nontrivial expense of diagnostic accuracy.

Even among expert radiologists, clinically substantial errors are made in 3-6\% of studies\cite{Robinson1999,Brady2012}, with minor errors seen in 30\% \cite{Bruno2015}.  Accurate diagnosis of some entities is particularly challenging: early lung cancer for example is missed in 19-54\% of cases, with similar sensitivity figures described for pneumothorax and rib fracture detection. 
The likelihood for major diagnostic errors is directly correlated with both shift length and volume of examinations being read\cite{Hanna2017}, a reminder that diagnostic accuracy varies substantially even at different times of the day for a given radiologist. 

Hence there exists an immense unmet need and opportunity to provide immediate, consistent and expert-level insight into every CXR.  In the present work we describe a novel methodology employed in this endeavor and we present the results achieved using a robust method of clinical validation. 
 
\begin{figure}[t]
	\begin{center}
        \includegraphics[width=\textwidth]{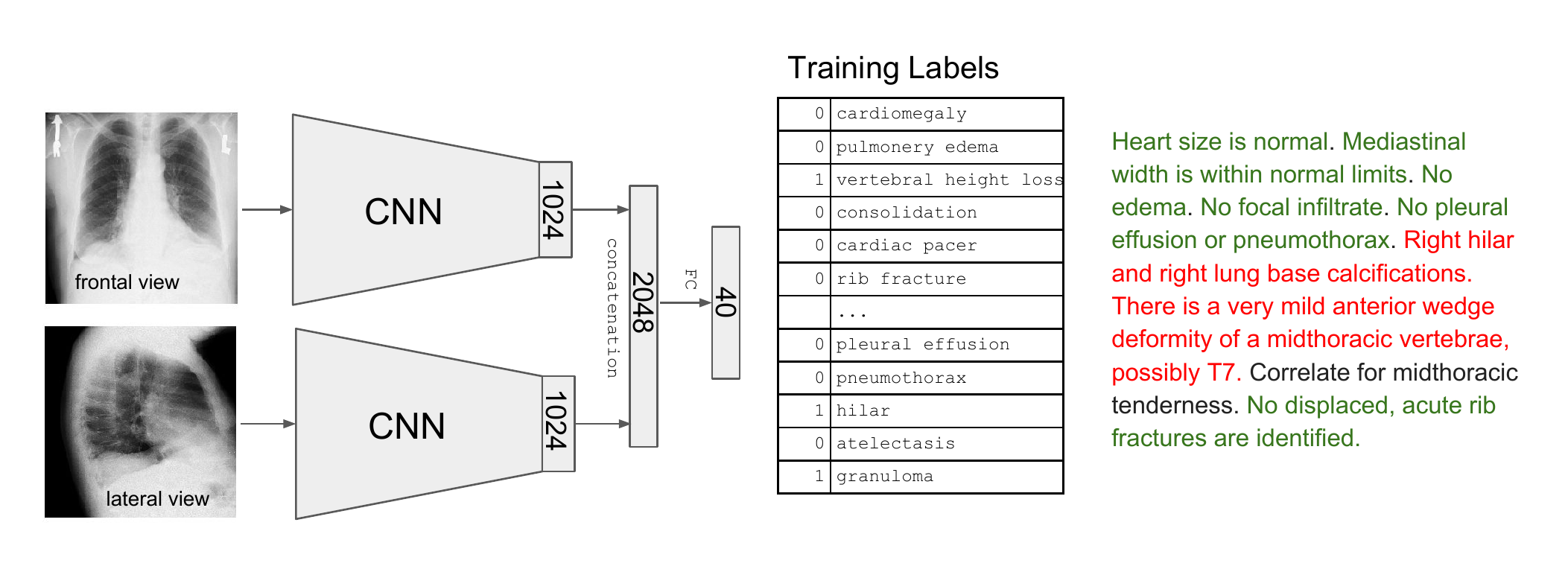}
	\end{center}
	\caption{TextRay Model Illustration. Frontal (PA) and lateral view images each go through a separate CNN. A fully-connected layer is applied on their concatenated feature vectors and emits the confidence for each finding. Training labels were extracted by analyzing the report sentences. Negative (green) and positive (red) sentences identified. Findings in positive sentences receive a positive training label. Negative or unmentioned findings receive a negative label.}
    \label{figure:model}
    
\end{figure}

\section{Material and Methods}
\subsubsection{Data}
All Patient Health Information (PHI) was removed from the data prior to acquisition in compliance with HIPAA  standards.  We utilized a dataset of 2.1 million CXRs with their respective diagnostic reports.    
All postero-anterior (PA) CXR films of individuals aged 18 and above were procured. Corresponding lateral views were present in 85\% of the CXR examinations and were included in the study data.  

\subsubsection{Textual Analysis}
A standardization process was employed whereby all CXR reports were reduced to a set of distinct canonical labels.  First, a sentence boundary detection algorithm was applied to the 2.1M reports, yielding a pool of 827k unique sentences.  Three expert radiologists and two medical students categorized the most occurring sentences with respect to their pertinence to CXR images.

Three categories emerged: sentences that report the presence or absence of a finding, for example \textit{"the heart is enlarged"}, or \textit{"normal cardiac shadow"}, and could be used as labels;
neutral sentences, which referenced information not derived from or inherently related to the image itself, for example: "\textit{84 year old man with cough}", "\textit{lung nodule follow up}", or "\textit{comparison made to CT chest}".  

A third category of sentences could render the study unreliable for training due to ambiguity regarding the relationship of the text to the image, for example "\textit{no change in the appearance of the chest since yesterday}". 

After filtering out neutral and negative sentences using a few hand-crafted regular expressions, it was possible to fully cover 826k reports using just the 20k most prevalent positive sentences. The same expert radiologists reviewed each of these sentences and mapped them to an initial ontology of 60 findings which covered 99.99\% of all positive sentence volume.  

In making the final ontology, we focused on visual findings rather than clinical interpretations or diagnoses. We chose to merge some categories: osteoporosis was merged into \textit{osteopenia}, twisted and uncoiled aorta into \textit{abnormal aorta}, and bronchial markings into \textit{interstitial markings}, since it is often impossible to differentiate these based on the image alone. Although visually distinct, all tubes and venous lines were consolidated into two respective categories. The resulting 40 categories are presented in Table \ref{table:training}.

\begin{table}
\caption{Number of studies with each finding in our data. 596k (62\%) of the total 959k studies had no reported findings.}
\begin{center}
\begin{tabular}{r|l|r|l ||| r|l|r|l}
\hline
\textbf{\#}  &  \textbf{finding} &  \textbf{total} &  \textbf{\%} &  \textbf{\#} &  \textbf{finding} &  \textbf{total} &  \textbf{\%} \\
\hline
    1 &               abnormal aorta & 15,932 &    1.66 &      21 &                        mass &     633 &      0.07 \\
    2 &         aortic calcification & 11,508 &    1.20 &      22 &        mediastinal widening &   1,639 &      0.17 \\
    3 &             artificial valve &  5,847 &    0.61 &      23 &              much bowel gas &     441 &      0.05 \\
    4 &                  atelectasis &  5,492 &    0.57 &      24 &                      nodule &     553 &      0.06 \\
    5 &    bronchial wall thickening &  2,773 &    0.29 &      25 &          orthopedic surgery &     717 &      0.07 \\
    6 &                cardiac pacer & 17,378 &    1.81 &      26 &                  osteopenia &   5,585 &      0.58 \\
    7 &                 cardiomegaly & 95,137 &    9.92 &      27 &            pleural effusion &  16,688 &      1.74 \\
    8 &                 central line &  3,802 &    0.40 &      28 &          pleural thickening &   8,164 &      0.85 \\
    9 &                consolidation & 34,260 &    3.57 &      29 &                pneumothorax &     741 &      0.08 \\
   10 &  costophrenic angle blunting & 13,673 &    1.43 &      30 &             pulmonary edema &   8,637 &      0.90 \\
   11 &         degenerative changes & 18,545 &    1.93 &      31 &                rib fracture &   4,607 &      0.48 \\
   12 &           elevated diaphragm & 21,913 &    2.28 &      32 &                   scoliosis &   4,907 &      0.51 \\
   13 &             fibrotic changes & 11,027 &    1.15 &      33 &  soft tissue calcifications &   1,086 &      0.11 \\
   14 &                     fracture &    526 &    0.05 &      34 &            sternotomy wires &  45,002 &      4.69 \\
   15 &                    granuloma &  1,475 &    0.15 &      35 &        surgical clips noted &   8,147 &      0.85 \\
   16 &             hernia diaphragm &  8,892 &    0.93 &      36 &       thickening of fissure &   1,714 &      0.18 \\
   17 &                        hilar prominence & 10,407 &    1.08 &      37 &           trachea deviation &     601 &      0.06 \\
   18 &               hyperinflation & 37,319 &    3.89 &      38 &                  transplant &   5,180 &      0.54 \\
   19 &        interstitial markings & 97,703 &   10.18 &      39 &                        tube &   2,025 &      0.21 \\
   20 &                     kyphosis &  5,531 &    0.58 &      40 &       vertebral height loss &   1,212 &      0.13 \\
\hline
\end{tabular}
\end{center}
\label{table:training}
\end{table}

\vspace{-1cm}
\subsubsection{Training Set Generation}
On completion of sentence labeling, we set out to design the appropriate training set. A conservative approach would only include studies whose report sentences were \textit{fully-covered}, i.e. every potentially positive sentence in them was manually reviewed and mapped to a finding.  A more permissive \textit{any-hit} approach would include any study with a recognized positive sentence in its report, ignoring other unrecognized sentences, with the risk that some of them also mention abnormalities that would be mislabeled as negatives.  

The \textit{fully-covered} approach yielded 596k normal studies (no positive findings), and 230k abnormal studies. The \textit{any-hit} approach, while noisier, added 58\% more abnormal studies, for a total of 363k. Hence our final training set had 826k studies in the \textit{fully-covered} approach, and 959k studies in the \textit{any-hit} approach. 

Additionally, many radiologists will omit mention of normal structures in favor of brevity, thereby implying a negative label. This bias extends to many studies in which even mildly abnormal or senescent changes are omitted. For example, the same CXR may produce a single-line report of "\textit{No acute disease}" by one radiologist and descriptions of cardiomegaly, and degenerative changes by another radiologist. Inherently, this omission bias introduces noise into the labeling process, particularly for findings which are not deemed critical, even in the more conservative \textit{fully-covered} training set.

We decided to compare both approaches, and took the larger \textit{any-hit} training set as our baseline. To the best of our knowledge, this is the largest training set ever assembled for chest X-ray, both in terms of the number of studies and the number of labels (see Table \ref{table:training} for its composition). We partitioned the training set into \textit{training}, \textit{validation}, and \textit{testing} (80\%/10\%/10\% respectively), based on the (anonymized) patient identity.  From the 10\% of studies designated as \textit{validation} we compiled a validation set of size 994 with at least 25 positives from each finding. We picked the model with lowest validation loss.

\subsection{Model}
Our model, called TextRay, is illustrated in Fig. \ref{figure:model}. We start by applying a CNN (DenseNet121\cite{densenet}) on the Lateral and PA views (separately).  We removed the last fully connected layer from each CNN and concatenated their outputs (just after the average pooling layer). We then applied our own fully-connected layer resulting in $K=40$ outputs, one for each finding, followed by a sigmoid activation. Hence, our model treats each study as a “bag of findings”, reporting the confidence for each one. We used the mean of the binary cross-entropy losses as our main loss function:
\[
  loss=\frac{1}{K}\sum_{k=1}^K y_k\log(p_k)+(1-y_k)\log(1-p_k)
\]
where $p_k$ is the value of the $k$-th output unit and $y_k$ is the binary label for the $k$-th finding.

Our model receives two inputs of size 299x299. When lateral view was unavailable, we fed the network with random noise instead. Each X-Ray image (up to 3000x3000 pixels in raw format) was zero-mean-normalized, rescaled to a size of $330(1+a)$ x $330(1+b)$, and rotated c degrees. A random patch of 299x299 was taken as input. For training augmentation, we sampled $a, b$ uniformly from $\pm0.09$ and $c$ from $\pm9$, randomly flipping each image horizontally. For balance, we replaced the PA view with random noise in 5\% of the samples. For test we used $a=b=c=0$ and took the central patch as input, without flipping. 

We trained on two 1080Ti GPUs, putting each CNN on a different GPU. We used the built-in Keras 2.1.3 implementation of DenseNet121 over Tensorflow 1.4. We used the Adam optimizer with Keras’ default parameters, and a batch size of 32. We sorted the studies in two queues, normals and abnormals, and filled each batch with 95\% abnormal studies on average.  An epoch was defined as 150 batches. We started with a learning rate of 0.001 and multiplied it by 0.75 if validation loss hadn't improved for 30 epochs. We trained for 2000 epochs. 

\subsection{Evaluation Sets}

We chose 12 of the 40 findings and prepared evaluation sets for them, using studies from the \textit{test} partition. Most sets focused on a single finding except \textit{cardiomegaly}, \textit{hilar prominance}, and \textit{pulmonary edema}, which  were lumped together as they are commonly seen in the setting of congestive heart failure. In each set, the studies were derived from two pools: \textit{pos-pool} are studies that the reports indicated as positive for that finding. These studies were obtained by a manual textual search for terms indicative for each finding, independently of our sentence-tagging operation; \textit{neg-pool} are randomly sampled studies, which are mostly negative for any finding (see Table \ref{table:testing} for the sets composition).

Each set was evaluated by three expert radiologists.
In each set, the radiologist reviewed the shuffled studies and indicated the presence or absence of the relevant finding, using a web-based software operated on a desktop. The radiologists were shown both PA and Lateral view in their original resolutions.

We considered the report as a fourth expert opinion. To measure the accuracy of the label-extraction process, we cross referenced the report opinion with the training set labels. The positive labels in the training set were accurately mentioned the report; frequently, positive findings mentioned in the reports were mislabeled as negatives, (see Table S\ref{table:noise}) as would be expected in the \textit{any-hit} training set, but this was also observed to lesser degree even in the \textit{fully-covered} set.

\begin{table}
\caption{Evaluation Sets. The number of studies taken from the \textit{pos-pool} (finding is positive in report) and \textit{neg-pool} (random sample) are indicated, along with the average agreement rate (AAR) of the 3 radiologists (rads) assigned to each set vs.\ the report.
The AAR between our model and the rads (column \textit{textray}) is compared against the AAR between any radiologist and the other rads (\textit{avg.\ rad.}) . Confidence intervals are computed over the difference ($\Delta =\mbox{textray}-\mbox{avg.\ rad.}$). }
\begin{center}
\begin{tabular}
{l|c|c||c|c|c | l}
\hline &
\multicolumn{2}{|c||}{pool} & 
\multicolumn{3}{|c|}{avg.\ agreement w/ rads} &  
\multicolumn{1}{c}{$\Delta$ (CI)} \\
\cline{2-6} 
\textbf{finding} &  pos &  neg &  report & avg. rad & textray &  textray vs.\ rads\\
\hline
\hline
\textbf{pulmonary edema      } &       128 &       482 &             0.613 &               0.639 &              0.730 &            +0.09 (0.07, 0.11) \\
\textbf{elevated diaphragm   } &       202 &        77 &             0.731 &               0.675 &              0.754 &            +0.08 (0.05, 0.10) \\
\textbf{abnormal aorta       } &       198 &        80 &             0.736 &               0.693 &              0.771 &            +0.08 (0.05, 0.11) \\
\textbf{hyperinflation       } &        95 &        80 &             0.678 &               0.619 &              0.657 &           +0.04 (-0.02, 0.10) \\
\textbf{vertebral height loss} &       126 &        55 &             0.781 &               0.742 &              0.757 &           +0.02 (-0.02, 0.06) \\
\textbf{atelectasis          } &       201 &        78 &             0.778 &               0.756 &              0.767 &           +0.01 (-0.03, 0.04) \\
\textbf{cardiomegaly         } &       238 &       372 &             0.755 &               0.861 &              0.866 &           +0.01 (-0.02, 0.03) \\
\textbf{pleural effusion     } &       207 &        73 &             0.905 &               0.893 &              0.896 &           +0.00 (-0.02, 0.03) \\
\textbf{consolidation              } &       194 &        78 &             0.690 &               0.730 &              0.707 &           -0.02 (-0.07, 0.02) \\
\textbf{pneumothorax         } &       111 &       124 &             0.830 &               0.855 &              0.823 &           -0.03 (-0.08, 0.01) \\
\textbf{rib fracture         } &       183 &        76 &             0.683 &               0.799 &              0.745 &          -0.05 (-0.10, -0.01) \\
\textbf{hilar prominence     } &       184 &       426 &             0.552 &               0.797 &              0.736 &          -0.06 (-0.09, -0.03) \\
\hline
\end{tabular}
\end{center}
\label{table:testing}
\end{table}

\section{Results}
We performed pairwise analysis of the radiologist agreement following the procedure in \cite{CheXNet}, except we used the agreement rate between two taggers (e.g. accuracy) instead of the F1 score, because (a) it also measures agreement on the negatives; and (b) it is easier to interpret.  The \textit{average agreement rate} (AAR) for a radiologist (or a model) is the average of the agreement rates achieved against the other two (three for a model) radiologists. The \textit{avg. radiologist rate} is the mean of the three radiologists' AARs. We used the bootstrap method ($n=10000$) to obtain 95\% confidence intervals over the difference between TextRay and the average radiologist agreement rates. As TextRay's threshold for each finding, we used the one that maximized the AAR on the validation set.  

Table \ref{table:testing} shows that TextRay is on par with human radiologists (within the 95\% CI) on 10 out of 12 findings, with the exception of \textit{rib fracture} and \textit{hilar prominence}. On some findings (\textit{elevated diaphragm}, \textit{abnormal aorta}, and \textit{pulmonary edema}), radiologists agree significantly more with our algorithm than with each other (e.g. the CI does not include 0). Table \ref{table:testing} also shows the average agreement of the radiologists with the report. Here as well, this agreement is often higher than the average agreement among the radiologists themselves. This provides evidence that the noise added by using the reports as labels is no larger than the noise added by training a radiologist to do the tagging.

\begin{figure}
	\begin{center}
        \includegraphics[width=\textwidth]{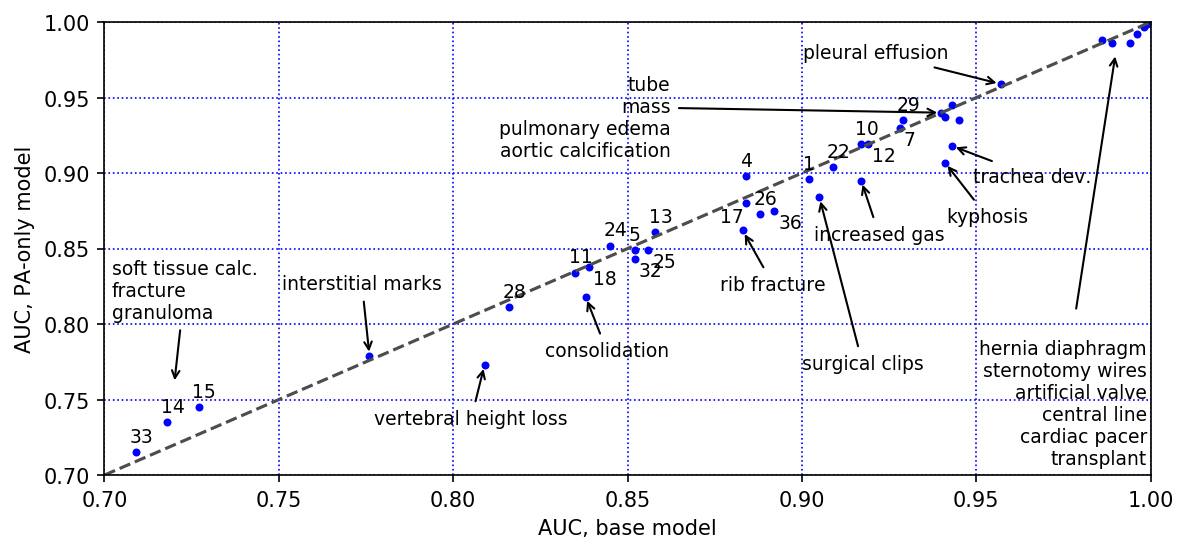}
	\end{center}
	\caption{Area under the ROC curve (AUC) of our base model vs. the PA-only variant over 40 chest X-ray findings. The numbers refer to the index of Table \ref{table:training}. A cluster of labels should be mapped left-to-right.}
    \label{figure:aucs}
\end{figure}

\begin{figure}
	\begin{center}
        \includegraphics[width=\textwidth]{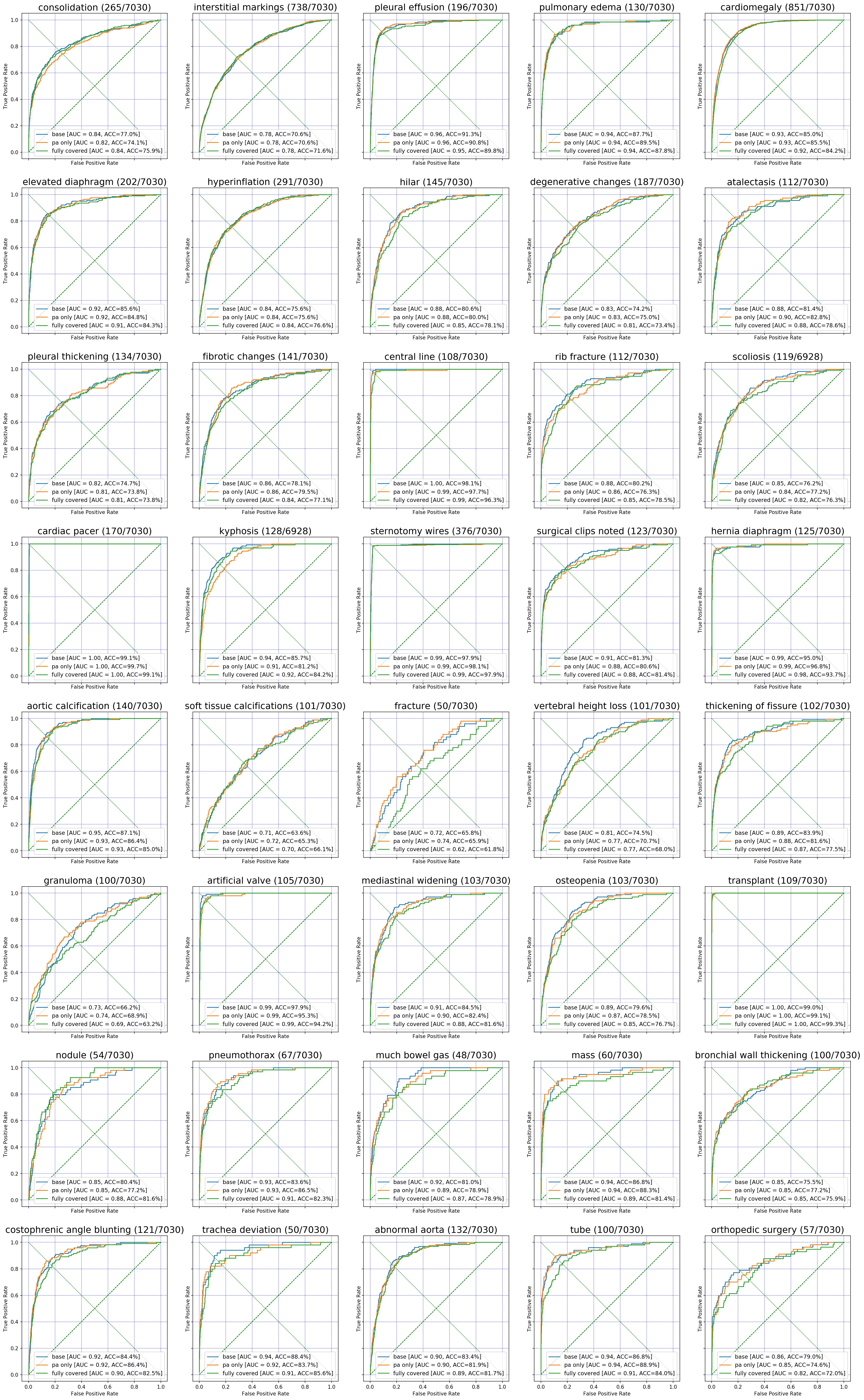}
	\end{center}
	\caption{ROC plots of our base model and its two variants on 40 chest X-ray findings. The title for each plot indicates the number positives in the test set of 7,030 studies. Indicated on each plot is the AUC and the accuracy when sensitivity=specificity.}
    \label{figure:rocs}
\end{figure}

Using our text-based labels as ground-truth, TextRay's performance was then tested over all 40 findings. To create the test set, a random sample of 5,000 studies was chosen from the \textit{test} partition. Then, more studies were added from the partition until each finding had at least 100 positive cases, for a total of 7,030 studies. The ROC plots are shown in supp. Fig. \ref{figure:rocs}, with their AUCs ranging between 0.7 and 1.0 (average 0.892). At the top of the chart, artificial objects (i.e. pacers, lines, tubes, wires, and implants) are detected with AUCs approaching 1.0, much better than all diseases. 

Fig. \ref{figure:aucs} shows the area under the ROC curve (AUC) achieved by our model compared to a variant that was trained only with the PA view of each study (the approach used in \cite{CheXNet,Chestxray8}). We see that in most findings, the performance is similar, but \textit{vertebral height loss}, \textit{consolidation}, \textit{rib fracture}, and \textit{kyphosis} stand out as findings in which the lateral view improved detection. These findings are expected from a clinical radiographic perspective. 

\begin{figure}
	\begin{center}
        \includegraphics[width=\textwidth]{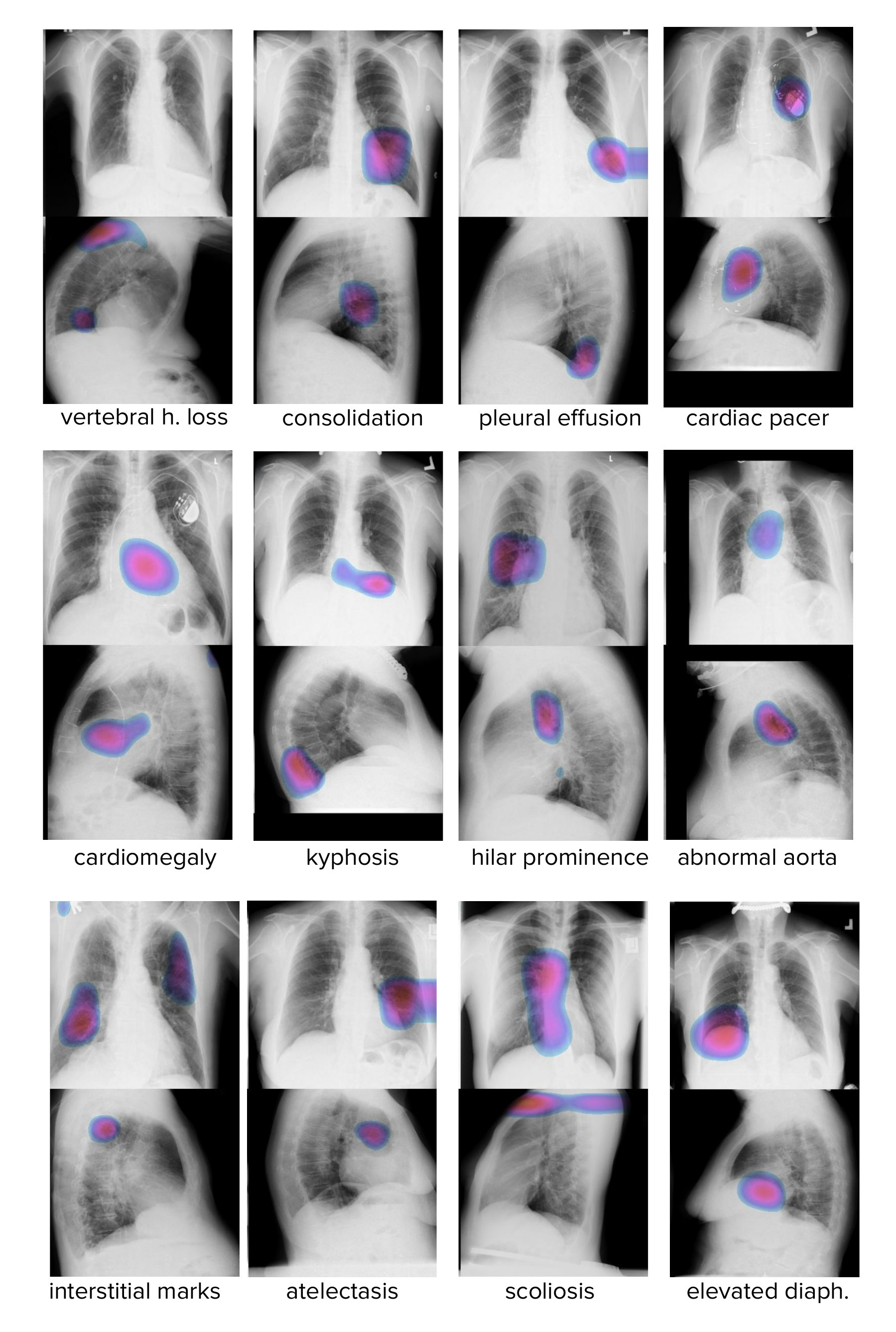}
	\end{center}
	\caption{Heat maps for 12 positive findings on a selected set of studies. For each finding, a heat map is generated as a linear combination over the 1024 feature maps calculated for each view. The weight given to feature-map $i$ when generating a heat map for finding $j$ is $W_{ij}$, where $W$ are the weights of the last fully-connected layer of the network.}
    \label{figure:heatmaps}
\end{figure}

For comparison, we also trained a variant of TextRay with the \textit{fully-covered} training set, but it achieved significantly lower results in almost all findings (see supp. Table \ref{table:aucs}), suggesting that the additional abnormal studies in the \textit{any-hit} set, more than compensated for the higher label noise. Finally, we draw heat maps based on the procedure presented in \cite{Chestxray8}, and present them in Supp. Fig. \ref{figure:heatmaps}.

\section{Discussion}
The extraction of labels from full CXR reports has been recognized as essential for efficient and robust CNN training on large datasets.  Shin et al.\cite{Shin_2016_CVPR} extracted labels from the 3,955 CXR reports in the OpenI dataset, using the MeSH system\cite{Demner-Fushman2015AnnotationRetrieval}.  
The ChestX-ray14 dataset released by Wang et al.\cite{Chestxray8} contains 112k PA images loosely labeled using a combination of NLP and hand-crafted rules. Rajpurkar et al.'s \cite{CheXNet} team of four radiologists reported a high degree of disagreement with the provided ChestX-ray14 labels in general, although they demonstrate the ability to achieve expert-level prediction for the presence of pneumonia after training upon a DenseNet121 CNN.

Utilizing several public datasets with image labels and reports provided, Jing et al.\cite{Jing2017OnReports} built a system that can generate a natural appearing radiology report using a hierarchical RNN. The high-level RNN generates sentence embeddings that seed low-level RNNs that produce the words of each sentence. As part of their report generation, they also produce tags representing the clinical finding present in the image. Interestingly, the model trained using these tags and the text of the reports did not predict the tags better than the model that was trained just using the tags. The ultimate accuracy of the system however remains poorly defined due to lack of clinical radiologic validation. 
   
To the best of our knowledge, the present study is the first to utilize extensive radiology expertise for both multi-label generation and visual validation of algorithmic results.  Study labels were generated bottom-up via ontology-based methodology which was rooted in the text rather than pre-existing categories or tags (i.e. MeSH). We trained upon the largest dataset of CXRs described to date, achieving results on twelve distinct visual findings which are on par with inter-radiologist agreement and in some cases, better.   

\section{Conclusion}
In this work we attempt to broadly cover all findings radiologists usually report when reviewing a PA and Lateral chest X-ray. Since a relatively small set of sentences is heavily re-used in CXR reports, we were able to to generate organic labels for millions of reports by examining and indexing twenty thousand individual sentences. This massive amount of data allowed us to obtain radiology-level detection performance on various of findings using a single model, in essence distilling the insight of millions of radiographic interpretations into software code.  Application of a similar technique upon AP chest X-ray scans, musculoskeletal and abdominal radiographies is currently ongoing. 

%
%
%
%
%

%
%
\bibliography{references}

\newpage
\section*{\Huge{Supplementary Material}}

\begin{table}
\caption{Most frequent positive sentences and their occurrences in reports.}

\begin{center}
\begin{tabular}[b]{p{5cm} c || p{4cm} c}
\hline
  \textbf{sentence} & \textbf{\#reports} & \textbf{sentence} & \textbf{\#reports} \\ 
  \hline
  The heart is enlarged & 39,245 & 
  Twisted aorta & 6,771 \\ 
  The heart is widened & 20,270 & 
    Infiltrate? & 6,540\\
  Enlarged heart & 14,689   &  
  Increased lung volume & 6,494  \\
  Chronic bronchial changes & 9,515& 
  After sternotomy & 6,268 \\
  Enhanced interstitial markings in the lungs & 9,216  & 
  Interstitial changes in the lungs & 5,303\\ 
  Permanent cardiac pacer & 6,881  &  
  Hyperinflation & 5,064\\ 
      \hline
\end{tabular}
\end{center}
\label{table:sentences}
\end{table} 

\begin{figure}
	\begin{center}
        \includegraphics[width=\textwidth]{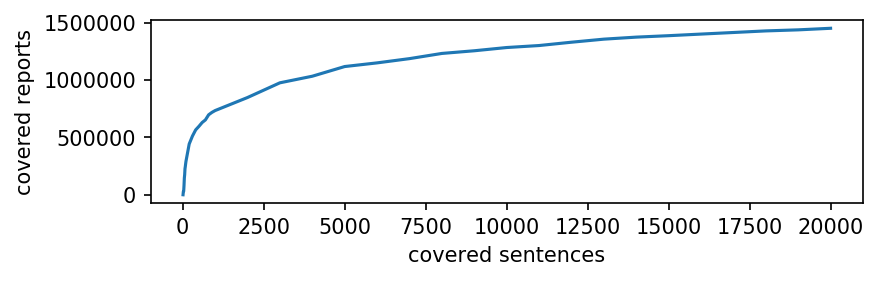}
	\end{center}
	\caption{Number of reports fully-covered by tagged sentences as a function of the number of tagged sentences (assuming we tag the most commons ones).}
\label{figure:coverage}
\end{figure}

\begin{table}
\begin{center}
\begin{tabular}{l|lllllll|p{1cm}}
\hline
& \multicolumn{7}{|c|}{AAR of each radiologist} & rad.\\
finding & A &  B &  C &  D &  E &  F &  G & average \\
\hline
\textbf{abnormal aorta       } &  0.75 &  0.72 &       &       &   0.6 &       &       &   0.69 \\
\textbf{atelectasis          } &       &       &  0.77 &       &  0.73 &  0.77 &       &   0.76 \\
\textbf{cardiomegaly         } &       &       &  0.85 &  0.86 &       &       &  0.87 &   0.86 \\
\textbf{elevated diaphragm   } &       &       &       &  0.67 &   0.6 &  0.75 &       &   0.68 \\
\textbf{hilar prominence     } &       &       &  0.79 &  0.79 &       &       &  0.82 &   0.80 \\
\textbf{hyperinflation       } &  0.65 &       &       &       &  0.69 &  0.52 &       &   0.62 \\
\textbf{consolidation        } &  0.77 &       &  0.72 &   0.7 &       &       &       &   0.73 \\
\textbf{pleural effusion     } &  0.91 &       &       &  0.88 &   0.9 &       &       &   0.89 \\
\textbf{pneumothorax         } &       &       &       &  0.87 &  0.84 &  0.85 &       &   0.86 \\
\textbf{pulmonary edema      } &       &       &  0.71 &  0.61 &       &       &   0.6 &   0.64 \\
\textbf{rib fracture         } &       &       &       &   0.8 &  0.79 &  0.81 &       &   0.80 \\
\textbf{vertebral height loss} &  0.75 &   0.8 &       &       &       &  0.68 &       &   0.74 \\
\hline
\end{tabular}
\end{center}
\label{table:taggers}
\caption{Evaluation sets and their assigned taggers (marked with letters A-G). The taggers A-G are attending radiologists with 40, 6, 5, 5, 2, 2, and 2 years of experience, respectively.  The numbers indicate the average agreement rate (AAR) of each radiologist vs.\ the other two radiologists in the set.}
\end{table}

\begin{table}
\begin{center}
\begin{tabular}{l|cc|cc}
\hline
 & \multicolumn{2}{|c|}{\% pos studies included}  
 & \multicolumn{2}{|c}{\% pos findings correctly labeled} \\
 \cline{2-5}
finding &         fully-covered &   any-hit &    fully-covered &   any-hit \\
\hline
\textbf{abnormal aorta       } &      44.9 &          81.8 &   97.8 &     87.7 \\
\textbf{atelectasis          } &      13.9 &          64.2 &   78.6 &     47.3 \\
\textbf{cardiomegaly         } &      55.0 &          95.0 &  100.0 &     99.1 \\
\textbf{elevated diaphragm   } &      42.1 &          76.7 &   95.3 &     81.9 \\
\textbf{hilar prominence     } &      43.5 &          84.2 &   96.2 &     86.5 \\
\textbf{hyperinflation       } &      56.8 &          85.3 &   90.7 &     91.4 \\
\textbf{consolidation              } &      24.2 &          50.5 &   87.2 &     58.2 \\
\textbf{pleural effusion     } &      34.3 &          80.7 &   94.4 &     70.1 \\
\textbf{pneumothorax         } &      19.8 &          56.8 &   50.0 &     39.7 \\
\textbf{pulmonary edema      } &      57.0 &          93.8 &   86.3 &     75.0 \\
\textbf{rib fracture         } &      34.4 &          63.4 &   82.5 &     61.2 \\
\textbf{vertebral height loss} &      20.6 &          64.3 &   73.1 &     39.5 \\
\hline
\end{tabular}
\end{center}
\label{table:noise}
\caption{Estimation of label noise in two training sets. The \textit{fully-covered} training set only includes a study if all its potentially positive sentences were parsed and mapped to their respective findings. The \textit{any-hit} training set includes a study if at least one sentence was parsed and mapped to a finding, even if the rest of the positive sentences were not parsed and their findings are unknown.  The \textit{any-hit} training set includes more positive studies for every finding (left 2 columns), but a larger portion of those positive findings is erroneously labeled as negative (right 2 columns).}
\end{table}

\begin{table}
\caption{Model performance per finding, measured in AUC.  PA: model trained only with PA view. FC: model trained only on studies with \textit{fully-covered} reports.}\begin{center}
\begin{tabular}{r|l|c|c|c||r|l|c|c|c}
\hline
 \# &  finding &  base  &  PA &  FC &  \# &                   finding (cont.) &  base &  PA &  FC \\
\hline
    1 &         abnormal aorta &  \textbf{0.902} &           0.896 &           0.887 &       21 &                   mass &  \textbf{0.941} &           0.937 &           0.894 \\
     2 &   aortic calcification &  \textbf{0.945} &           0.935 &           0.930 &       22 &   mediastinal widening &  \textbf{0.909} &           0.904 &           0.885 \\
     3 &       artificial valve &  \textbf{0.994} &           0.986 &           0.989 &       23 &         increased bowel gas &  \textbf{0.917} &           0.895 &           0.867 \\
     4 &            atelectasis &           0.884 &  \textbf{0.898} &           0.877 &       24 &                 nodule &           0.845 &           0.852 &  \textbf{0.882} \\
     5 &   bronchial wall thick &  \textbf{0.852} &           0.849 &           0.852 &       25 &     orthopedic surgery &  \textbf{0.856} &           0.849 &           0.817 \\
     6 &          cardiac pacer &  \textbf{0.998} &           0.997 &           0.997 &       26 &             osteopenia &  \textbf{0.888} &           0.873 &           0.846 \\
     7 &           cardiomegaly &           0.928 &  \textbf{0.930} &           0.918 &       27 &       pleural effusion &           0.957 &  \textbf{0.959} &           0.949 \\
     8 &           central line &  \textbf{0.996} &           0.992 &           0.990 &       28 &     pleural thickening &  \textbf{0.816} &           0.811 &           0.811 \\
     9 &          consolidation &  \textbf{0.838} &           0.818 &           0.838 &       29 &           pneumothorax &           0.929 &  \textbf{0.935} &           0.910 \\
    10 &  costoph. angle blunt. &           0.917 &  \textbf{0.919} &           0.896 &       30 &        pulmonary edema &           0.943 &  \textbf{0.945} &           0.944 \\
    11 &   degenerative changes &  \textbf{0.835} &           0.834 &           0.812 &       31 &           rib fracture &  \textbf{0.883} &           0.862 &           0.855 \\
    12 &     elevated diaphragm &  \textbf{0.919} &           0.919 &           0.908 &       32 &              scoliosis &  \textbf{0.852} &           0.843 &           0.824 \\
    13 &       fibrotic changes &           0.858 &  \textbf{0.861} &           0.839 &       33 &      soft tissue calc. &           0.709 &  \textbf{0.715} &           0.703 \\
    14 &               fracture &           0.718 &  \textbf{0.735} &           0.616 &       34 &       sternotomy wires &  \textbf{0.989} &           0.986 &           0.988 \\
    15 &              granuloma &           0.727 &  \textbf{0.745} &           0.689 &       35 &   surgical clips noted &  \textbf{0.905} &           0.884 &           0.883 \\
    16 &       hernia diaphragm &           0.986 &  \textbf{0.988} &           0.984 &       36 &  thickening of fissure &  \textbf{0.892} &           0.875 &           0.870 \\
    17 &                  hilar prominence &  \textbf{0.884} &           0.880 &           0.855 &       37 &      trachea deviation &  \textbf{0.943} &           0.918 &           0.908 \\
    18 &         hyperinflation &           0.839 &           0.838 &  \textbf{0.844} &       38 &             transplant &  \textbf{0.999} &           0.999 &           0.999 \\
    19 &  interstitial markings &           0.776 &           0.779 &  \textbf{0.781} &       39 &                   tube &  \textbf{0.940} &           0.940 &           0.911 \\
    20 &               kyphosis &  \textbf{0.941} &           0.907 &           0.925 &       40 &  vertebral height loss &  \textbf{0.809} &           0.773 &           0.766 \\
\hline
\end{tabular}
\end{center}
\label{table:aucs}
\end{table}

\end{document}